\documentclass[runningheads]{llncs}

\usepackage{eccv}

\usepackage{eccvabbrv}

\usepackage{graphicx}
\usepackage{booktabs}
\usepackage{multirow, makecell}
\usepackage{caption}
\usepackage[accsupp]{axessibility}  

\usepackage{hyperref}

\usepackage{orcidlink}

\usepackage{comment}
\usepackage{algorithm,algpseudocode}
\usepackage{mathtools}
\usepackage{colortbl}

\newcommand{\bz}{\mathbf{z}}

\newcommand{\bfd}{{\bf d}}

\newcommand{\bfc}{{\bf c}}
\newcommand{\defeq}{\coloneqq}

\newcommand{\myparagraph}[1]{\vspace{2pt}\noindent{\bf #1}}

\definecolor{yan}{rgb}{0.5,0.3,0.5}

\definecolor{ben}{rgb}{0.9,0.,0.5}

\definecolor{navyblue}{RGB}{191, 209, 229} 
\definecolor{light_yellow}{RGB}{255,243,194}
\definecolor{orange}{RGB}{255,200,100}

\newcommand{\best}{\cellcolor{orange}}
\newcommand{\second}{\cellcolor{light_yellow}}

\begin{document}

\title{\texorpdfstring{\raisebox{-0.0cm}{\includegraphics[height=0.4cm]{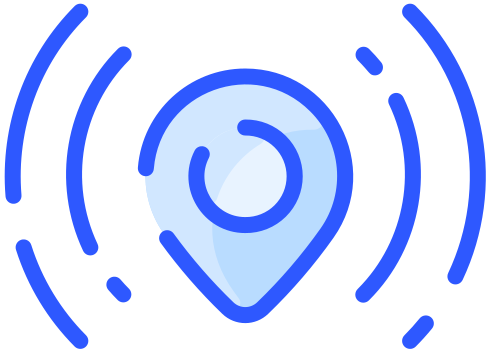}}\hspace{0.05cm}EchoScene: Indoor Scene Generation via Information Echo over Scene Graph Diffusion}{EchoScene}}

\titlerunning{EchoScene}

\author{Guangyao Zhai$^{1,3}$\orcidlink{0000-0002-6702-8302}
Evin P{\i}nar \"{O}rnek$^{1,3}$\orcidlink{0000-0003-1023-2852} \\
Dave Zhenyu Chen$^1$\orcidlink{0000-0002-3883-1905}
Ruotong Liao$^{2,3}$\orcidlink{0009-0003-1924-6502}
Yan Di$^1$\orcidlink{0000-0003-0671-8323}\hspace{1pt} \\
Nassir Navab$^1$\orcidlink{0000-0002-6032-5611}\hspace{1pt}
Federico Tombari$^{1,4}$\orcidlink{0000-0001-5598-5212}\hspace{1pt}
Benjamin Busam$^{1,3}$\orcidlink{0000-0002-0620-5774} 
}

\authorrunning{G. Zhai, E. P. \"{O}rnek, D. Z. Chen, R. Liao, Y. Di, N. Navab, et al.}

\institute{$^1$Technical University of Munich\hspace{7pt}
$^2$Ludwig Maximilian University of Munich\\
$^3$Munich Center for Machine Learning\hspace{10pt} 
$^4$Google\\
\url{guangyao.zhai@tum.de}\\[0.5ex]
\url{https://sites.google.com/view/echoscene}
}

\maketitle
\begin{abstract}
We present \emph{EchoScene}, an interactive and controllable generative model that generates 3D indoor scenes on scene graphs.
EchoScene leverages a dual-branch diffusion model that dynamically adapts to scene graphs.
Existing methods struggle to handle scene graphs due to varying numbers of nodes, multiple edge combinations, and manipulator-induced node-edge operations.
EchoScene overcomes this by associating each node with a denoising process and enables collaborative information exchange, enhancing controllable and consistent generation aware of global constraints.
This is achieved through an information echo scheme in both shape and layout branches. At every denoising step, all processes share their denoising data with an information exchange unit that combines these updates using graph convolution. The scheme ensures that the denoising processes are influenced by a holistic understanding of the scene graph, facilitating the generation of globally coherent scenes.
The resulting scenes can be manipulated during inference by editing the input scene graph and sampling the noise in the diffusion model.
Extensive experiments validate our approach, which maintains scene controllability and surpasses previous methods in generation fidelity. Moreover, the generated scenes are of high quality and thus directly compatible with off-the-shelf texture generation. Our code and models are open-sourced.
\end{abstract}

\begin{figure}
    \centering    \includegraphics[width=0.95\linewidth]{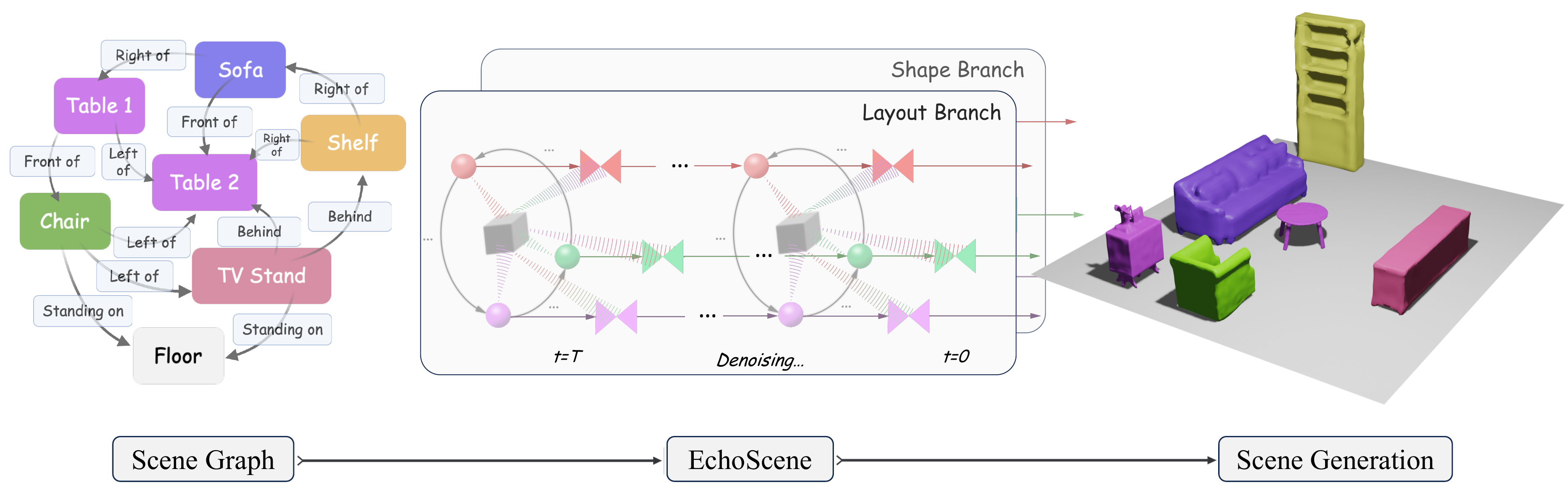}
    \caption{\textbf{EchoScene Schematic.} EchoScene uses a dual-branch diffusion model to generate 3D scenes from scene graphs. In both branches, each node is allocated a denoising process, and different processes are aware of global states through layout and shape "echoes" (waves in different colors) with an information exchange unit (grey block) along the denoising steps.}
    \label{fig:teaser}
\end{figure}
\section{Introduction}
\label{sec:intro}
Controllable Scene Generation (CSG) refers to synthesizing realistic 3D scenes according to input prompts while enabling specific entities within the scene to be user-interactive~\cite{chang2014learning,chang2015text,Simsar_2023_ICCV}. It has successfully been applied in robotics~\cite{zhai2023sg,mandlekar2023mimicgen}, Virtual Reality / Augmented Reality~\cite{bautista2022gaudi}, and autonomous driving~\cite{pronovost2024scenario,li2023drivingdiffusion}.
Recently, combining CSG with scene graph diffusion models has attracted significant research interest~\cite{commonscenes2023,tang2023diffuscene,lin2024instructscene}, since on the one hand, diffusion models empower more realistic and diverse 3D content generation~\cite{chen2023scenetex, chen2023text2tex, shi2024mvdream, poole2023dreamfusion}, on the other hand scene graphs capture the structure of the scene in a compact manner and facilitate intuitive user manipulation~\cite{graph2scene2021,commonscenes2023}.
Importantly, users can modify the input scene graph to dynamically change the generated scene.

Despite its significant progress so far, CSG with scene graph diffusion still suffers from two open challenges.
\textbf{First}, due to varying numbers of graph nodes and manipulator-induced node-edge operations, the input scene graphs dynamically describe global scene states, thus demanding adaptability from networks to accurately represent changing states.
This status includes nodes appearing or disappearing and edges altering during scene creation. 
\textbf{Second}, it is crucial yet difficult when encapsulating both fine-grained node classes and diverse edge combinations into a network to be aware of global constraints.
Existing solutions either simplify graphs by preserving only node sets, avoiding complex edge encoding in the diffusion while losing key structural relationships~\cite{tang2023diffuscene}, or convert nodes and edges into tokens for transformer-based denoiser~\cite{lin2024instructscene}, which is effective for the layout generation from simple graphs but not scalable for larger ones due to exponential token growth. For a graph with $P$ nodes and $Q$ types of edges, the maximum token count grows with $Q\cdot P!$, making such a strategy hard to model intricate relations and computationally infeasible for large graphs. 
CommonScenes~\cite{commonscenes2023} collapses graphs to triplets (subject, predicate, object) and employs graph convolution to aggregate features for each node, which are then conditioned on individual denoising processes for shape generation. While this approach adapts well to varying graph sizes, each denoising process is isolated from the other, resulting in insufficient awareness of global shape states and, thus, undesired inter-object inconsistency. Additionally, its reliance on VAE and GAN for layout generation complicates synchronized training.

To fully utilize the controllability of scene graphs for scene generation and manipulation, the feature aggregation of multiple edges to the respective nodes is essential.
In an attempt to flexibly encapsulate information from an indefinite number of nodes without losing global graph constraints, it can be helpful to allocate an individual denoising process per node and encourage information communication among all processes to achieve a common goal state.
Integrating both concepts, CSG becomes promising with complex scene graphs by adhering to graph constraints and manipulating object-level diffusion. 
With \emph{EchoScene} we move a step further and introduce a scene-generative framework equipped with dual-branch diffusion models capable of producing object shapes and scene layouts simultaneously (see Fig.~\ref{fig:pipeline} for the pipeline).
Each branch contains multiple denoising processes, with an equal number of nodes in the scene graph. The denoiser is weight-shared for all processes without introducing additional costs.
To consolidate global awareness in every denoising process, we introduce an \textit{information echo} scheme to exchange information at each denoising step in the respective branches, as illustrated in the middle of Fig.~\ref{fig:teaser}.
Here, an information echo in the graph diffusion happens when a node first sends the denoising data and other features to an information exchange unit, which echoes back the aggregated features to the node as the condition signal for its denoiser. 
More clearly, for a single denoising process, the echo route is: \{\textit{current denoising input} $\longrightarrow$ \textit{information exchange unit} $\longrightarrow$ \textit{denoising conditioner}\}.

In this design, EchoScene is able to generate and edit scenes from complex scene graphs with an indefinite number of nodes and edges. More importantly, the information echo scheme supports temporally global information exchange within both branches, thus continuously making the generation compliant with the scene graph description. For example, in the shape branch, shape echoes enable steady awareness of object appearance along the denoising steps for each generation process, bringing more inter-object consistency to the global style. Last but not least, the pure diffusion-based model dynamics enables synchronized training. EchoScene outperforms the previous state-of-the-art method on generation fidelity by a large margin and is more robust when handling graph manipulation. Moreover, we show that scenes generated by EchoScene are compatible with off-the-shelf texture generators, \textit{e.g.}, SceneTex~\cite{chen2023scenetex}, facilitating potential downstream applications.

The main contributions of this work can be summarized as follows:

\begin{enumerate}

    \item We present \emph{EchoScene}, a scene generation method with a dual-branch diffusion model on dynamic scene graphs, to simultaneously generate layouts and shapes with more controllability.
    
    \item We introduce an information echo scheme inside each branch of EchoScene that allows multiple denoising processes to exchange their denoising status among each other at each time step, bringing global awareness to each individual process.
    
    \item We show in the experiments that the proposed generative framework achieves more generation fidelity, more robustness in handling graph manipulation, and effectively handles inter-object style consistency.
    
\end{enumerate}

\section{Related work}
\label{sec:rewe}

\myparagraph{Semantic Scene Graph.}
Scene graphs have become a fundamental tool for semantic scene understanding, offering a structured and symbolic representation of scenes through nodes (objects) and edges (relationships)~\cite{image_retrieval_using_scene_graphs}. Their versatility spans various domains, including image manipulation, caption generation, and visual question answering, demonstrating their capacity to enrich 2D image generation~\cite{johnson2018image,dhamo2020,visual_genome}. Beyond 2D, scene graphs have been extended to 3D scene understanding \cite{3DSSG2020, 4D_OR}, dynamic modeling \cite{3d_dynamic_scene_graphs}, and notably, controllable scene synthesis \cite{Li2018grains}, showcasing their adaptability in representing complex spatial and semantic relations in multi-dimensional spaces~\cite{3DSSG2020,3d_dynamic_scene_graphs,Li2018grains,zhou2019scenegraphnet,ozsoy2021multimodal,commonscenes2023,kong2020semantic}. EchoScene proposes an indoor 3D scene generation method conditioned fully on semantic scene graphs coupled with a scene controlling ability.  

\myparagraph{Diffusion Models.}
Applications of diffusion models \cite{pmlr-v37-sohl-dickstein15} have witnessed rapid expansion, from generating intricate images to modeling complex distributions in various data types~\cite{NEURIPS2020_4c5bcfec,song2021scorebased}. Recent advances have focused on improving their flexibility and realism, with significant efforts to enhance conditional generation and refine the models' understanding of input conditions~\cite{meng2022sdedit,controlnet2023,kong2024eschernet,chefer2023attendandexcite,brooks2022instructpix2pix}. Score Distillation Sampling (SDS) was proposed to enable training Neural Radiance Fields through pretrained diffusion models \cite{poole2023dreamfusion}. A line of work addressed text-to-3D via SDS\cite{shi2024mvdream,Liu_2023_ICCV}, studied the compositionality of scenes via SDS\cite{lin2023componerf,cohen2023set, gao2023graphdreamer,Po2023Compositional3S}, and more recently via Gaussian Splatting \cite{zhou2024gala3d}. These methods can create realistic 3D assets from text via a small number of images; however, they are limited in terms of number of objects within a scene while capturing their relationships. Comparably, a set of works used diffusion models to learn 3D shapes explicitly \cite{cheng2023sdfusion, Liu2023MeshDiffusion}. Finally, diffusion models have recently been shown to be useful for graph generative processes \cite{vignac2022digress,verma2022varscene,kong2023autoregressive}. EchoScene draws from the recent advances in diffusion models, where both shape and layout generation are through graph diffusion processes.

\myparagraph{Controllability in Scene Synthesis.}
The quest for controllable scene synthesis has led to diverse methodologies, ranging from text descriptions to spatial layouts and probabilistic grammars for creating 3D scenes. Methods have explored generating scenes from images~\cite{chen2023scenedreamer}, texts~\cite{chang2015text}, and layouts~\cite{bahmani2023cc3d}, employing various strategies such as autoregressive models and deep priors to achieve controllable outputs~\cite{Ma2018language,Jyothi_2019_ICCV,Paschalidou2021NEURIPS}. Recent work has also seen graph-based conditioning emerge as a powerful approach for controlled scene synthesis, enabling more precise manipulation of scene elements and their relations~\cite{Li2018grains,zhou2019scenegraphnet,commonscenes2023,tang2023diffuscene,lin2024instructscene}. 
CommonScenes~\cite{commonscenes2023} came up with a scene graph conditioned scene generation method, which is based on a VAE and a GAN for capturing the layout and creating shapes via a latent diffusion model. Even though these methods have high quality in layout generation, their shape generation abilities are limited, as well as their manipulation quality still has room for improvement. EchoScene addresses these challenges through a diffusion-based layout generation with a novel information echo scheme.

\section{Preliminary}
\myparagraph{Scene Graph.}
\label{scenegraph}
The scene graph we use is semantic scene graph~\cite{chang2021comprehensive}, denoted as $\mathcal{G} = \left\{\mathcal{V}, \mathcal{E}\right\}$, which serves as a structured representation of a visual scene.
In such representation, $\mathcal{V} = \{v_{i}~|~i = {1, \ldots, N \} }$ refers to the set of object nodes, while $\mathcal{E} = \{e_{i \to j}~|~i,j = {1, \ldots, N}, i \neq j \}$ represents the set of directed edges connecting each pair of nodes $v_{i} \rightarrow v_{j}$. 
As structured in the Fig.~\ref{fig:pipeline}.A.1, each node $v_{i}$ and edge $e_{i\to j}$ can encompass various extensible attributes, \textit{e.g.}, object categorical information. 

\myparagraph{Scene Graph Convolution.}
\label{gcn}
The scene graph convolution we use in this work, originally from~\cite{johnson2018image} and applied in~\cite{Luo_2020_CVPR,graph2scene2021,commonscenes2023}, is the fundamental module processing semantic scene graphs, as known as triplet-GCN. A typical $K$-layer triplet-GCN module is:  
\begin{equation}
\begin{aligned}
&(\alpha_{v_{i}}^{(k)}, \beta_{e_{i\to j}}^{(k+1)}, \alpha_{v_{j}}^{(k)}) = h_1(\beta_{v_{i}}^{(k)}, \beta_{e_{i\to j}}^{(k)}, \beta_{v_{j}}^{(k)}), k = 0, \ldots, K - 1, \\
&\beta_{v_{i}}^{(k+1)} = \alpha_{v_{i}}^{(k)} + h_{2}\Big( \texttt{Avg}\big( \alpha_{{v}_{j}}^{(k)}~|~v_j \in N_{\mathcal{G}}(v_{i}) \big) \Big),
\end{aligned}
\label{eq:gcn}
\end{equation}
where $k$ represents an individual layer in triplet-GCN, and $h_1,h_2$ are MLP. $N_{\mathcal{G}}(v_{i})$ encompasses all nodes that are directly connected to $v_{i}$, and \texttt{Avg} denotes the process of average pooling. The initial attributes of nodes and edges, denoted by $(\beta_{v_{i}}^{(0)}, \beta_{e_{i\to j}}^{(0)}, \beta_{v_{j}}^{(0)})$, able to be customized for specific usages.

\myparagraph{Contextual Graph.} 
CLIP features are proven to bring strong inter-object information to the scene graph~\cite{commonscenes2023}. Thus, a semantic scene graph evolves to a contextual graph by using the readily available CLIP text encoder to infer the semantic information for each node and each triplet in the graph to provide semantic anchors ($p_i$, $p_{i \to j}$), and each node and edge have their own learnable vectors ($o_i$, $\tau_{i \to j}$). As shown in the Fig.~\ref{fig:pipeline}.A.1, the node and edge feature of a contextual graph is $v_i:=\{p_i, o_i\}$ and $e_{i \to j}:=\{p_{i \to j}, \tau_{i \to j}\}$, respectively.

\myparagraph{Conditional Diffusion Models.} 
Diffusion models learn to estimate a target distribution through a progressive Markov Process of length $T$ where a denoiser $\varepsilon_\theta$ is trained to gradually remove noise from diffused noisy versions $\bfd_{t}$ with $t \in \{ 1, \ldots, T\}$ of a data sample $\bfd$~\cite{NEURIPS2020_4c5bcfec,song2021scorebased}.
Latent Diffusion Models (LDMs)~\cite{ldm22} efficiently establish this process in latent space where a conditioner $\bfc$ can guide the generation process~\cite{cheng2023sdfusion}.
Learning the denoising is reached by minimization of the objective
\begin{equation}
\label{eqn:cldm}
    \mathcal{L}_{LDM} = 
    \mathbb{E}_{\bfd, \varepsilon\sim\mathcal{N}(0,1),t} 
    \left [ || \varepsilon - \varepsilon_\theta(\bfd_t, t, \bfc) ||_2^2 \right ].
\end{equation}
Once the denoiser $\varepsilon_\theta$ is learned, the inverse process can be established using Langevin dynamics~\cite{song2019generative}.
\section{Method}
\myparagraph{Overview.}
We present \emph{EchoScene}, a method that accomplishes scene generation through layout and shape generation from scene graphs. EchoScene evolves the contextual graph to the latent space utilizing an encoder and a manipulator based on triplet-GCN, as shown in Fig.~\ref{fig:pipeline}.A and as introduced in Sec.~\ref{evolve}. Then, as shown in Fig.~\ref{fig:pipeline}.B, it conditions latent nodes to layout branch (Fig.~\ref{fig:pipeline}.B.1) and shape branch (Fig.~\ref{fig:pipeline}.B.2) separately based on an information echo scheme defined in Sec.~\ref{echodiff}. In the layout branch, each diffusion process interacts with each other through layout echo to acquire global layout awareness at each denoising step, so that the final layout generation is compliant with the scene graph description (see Sec.~\ref{lb} and Fig.~\ref{fig:detail}.A). Similarly, in the shape branch, each diffusion process interacts with each other through shape echo to be aware of other shape appearances at each denoising step, so that the final generated shapes in the scene are consistent (see Sec.~\ref{sb} and Fig.~\ref{fig:detail}.B). Moreover, our generated scenes are compatible with off-the-shelf texture generators, \textit{e.g.}, SceneTex~\cite{chen2023scenetex}, making more photorealistic appearances.

\subsection{Graph Preprocessing}
\label{evolve}
\myparagraph{Graph Encoding.} 
 To make the layout and shape branches aware of the semantic and spatial information among the objects, we first encode the contextual graph to have latent relation embedding for each node by a triplet-GCN-based encoder $E_r$. 
In this case, we initialize the input embeddings of layer $0$ with the features from the contextual graph, thus, $(\beta_{v_{i}}^{(0)}, \beta_{e_{i\to j}}^{(0)}, \beta_{v_{j}}^{(0)})=(v_i, e_{i \to j}, v_j)$ in Eq.~\ref{eq:gcn}.  After the encoding, node features evolve to $\mathcal{V}_\mathcal{Z} = \{v_i^z~|~i = 1, \ldots, N\}$, where $v_i^z$ consists of an explicit semantic representation $v_i$ with an implicit embedding $z_i$ encapsulating relation information with other nodes.

\begin{figure*}[t]%
    \centering
    \includegraphics[width=0.95\linewidth]{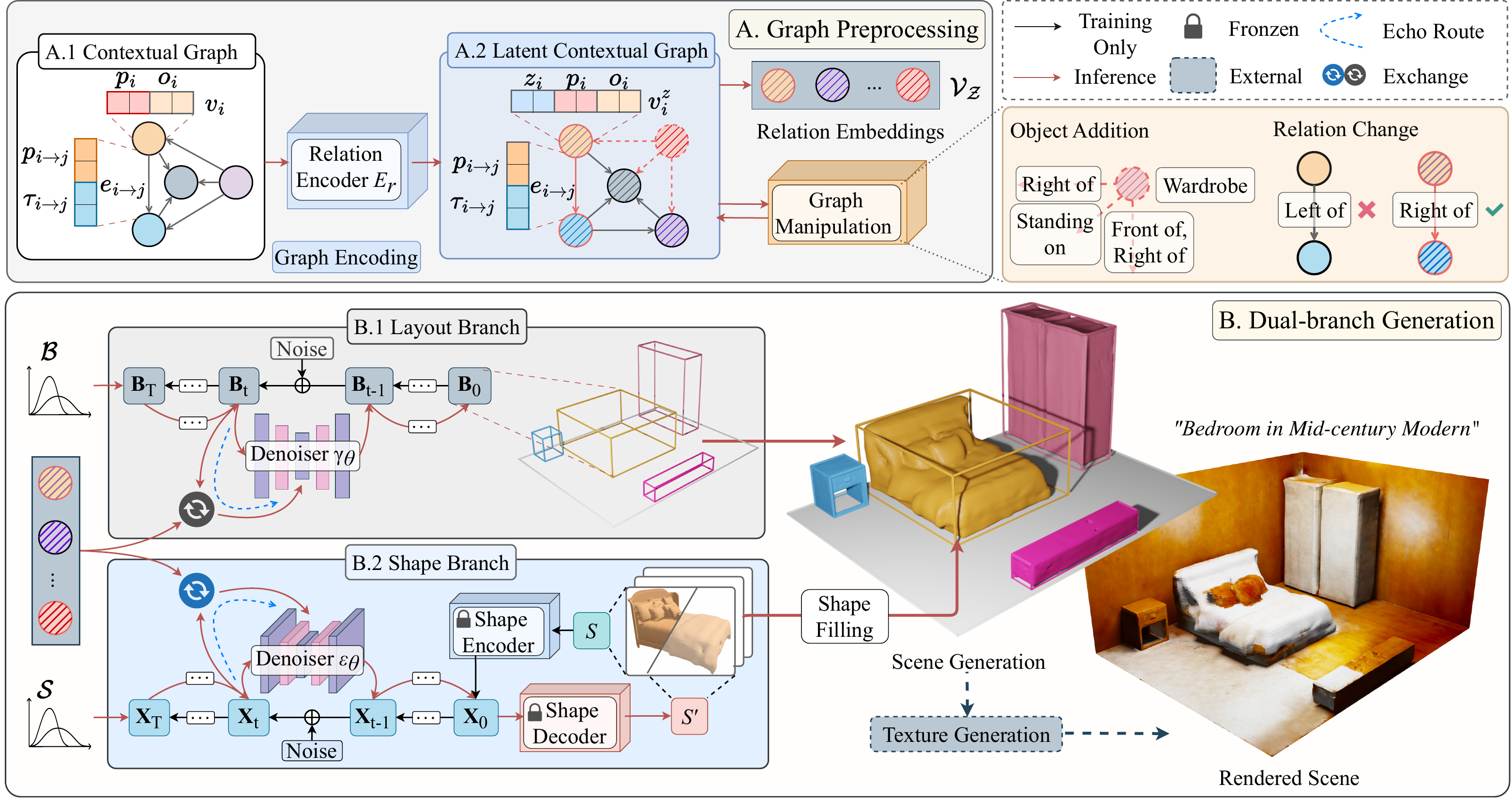}
    \caption{\textbf{Overview of EchoScene.} Our pipeline consists of graph preprocessing and two collaborative branches \textit{Layout Branch} and \textit{Shape Branch}. The details of two branches in one step are shown in Fig.~\ref{fig:detail}. During inference, EchoScene evolves the contextual graph to the latent space, where a manipulator optionally adjust the graph by editing nodes and edges. Then, EchoScene samples a random noise from Gaussian Distribution $\mathcal{B}$ and $\mathcal{S}$ for both branches conditioned on the latent graph to generate shapes and layouts. Finally, the generated shapes are populated into layouts to synthesize the scenes. Moreover, an external texture generator~\cite{chen2023scenetex} can be optionally utilized to provide a more photorealistic appearance.
    }
    \label{fig:pipeline}
\end{figure*}

\myparagraph{Graph Manipulation.}
The manipulation includes node addition and relation change, mimicking the user interaction. An example of manipulation is shown in Fig.~\ref{fig:pipeline}. We inherit the manipulation functionality of Graph-to-3D~\cite{graph2scene2021} and CommonScenes~\cite{commonscenes2023} by setting up a graph manipulator, another triplet-GCN module, to adjust the graph in the latent space, but with an orthogonal strategy during training. We illustrate more details in the Supplementary Materials.
After the manipulation, Node features in the scene graph are updated to  $\mathcal{V}_\mathcal{Z} = \{v_i^z~|~i = 1, \ldots, M, M \geq N$\}.
Subsequently, we send the updated graph to the diffusion-based layout branch and shape branch, where we set each node with an individual denoising process.
\subsection{Information Echo Scheme}
\label{echodiff}
Before introducing the layout and shape branch, we illustrate the core of the two branches--the information echo scheme in graph diffusion.



\myparagraph{Challenges from a Dynamic Graph.} 
While "Denoising Diffusion Probabilistic Models" (DDPM)~\cite{NEURIPS2020_4c5bcfec} has significantly advanced the field of generative models by producing realistic and diverse content, their application to dynamic data structures, such as scene graphs, is limited.
Scene graphs contain indefinite nodes and multiple edge combinations between nodes, describing compact and flexible relationships among objects. Some concurrent work based on DDPM~\cite{tang2023diffuscene,lin2024instructscene} to ours either models the object graph as a set or preserves it with a single edge between two nodes, where the representation ability of a scene graph degrades. Besides, they set a maximum number of tokens and model them in a single DDPM. As such, the local controllability of the generation is limited when users want to manipulate the specific elements. We believe a better solution is for each node in the graph to be allocated with an individual denoising process, jointly forming a diffusion model for specific tasks, which, in our case, are shape and layout generation. Thus, the content generation is fully controllable by node and edge manipulation. 

\myparagraph{Echo-Formed Condition.}
However, even though each denoising process theoretically realizes more controllability, it also brings isolation problems. As each generation proceeds individually, there is no awareness of scene content during the denoising steps, which makes the generation inconsistent with global constraints in the graph. To achieve both highly controllable and consistent generation at the same time, we propose to couple the inverse iterative conditional diffusion mechanism on graph node features with a recurrent intermediate message exchange strategy over the graph $\mathcal{G} = \left\{\mathcal{V}, \mathcal{E}\right\}$. In analogy to an echo in the real world, a node can send its information and receive it back along with information from other nodes during the steps of the inverse diffusion process. 

The cornerstone of echo is the introduction of an information exchange unit $U$ that enables dynamic and interactive diffusion processes among the elements of a dynamic graph. The incorporation of $U$ on the conditioner during the inverse diffusion process allows each element within the group to share and receive information.
At every denoising step, each diffusion process assembles the current denoising data $\bfd_t^i$ with other available node attributes, where we choose latent feature $v_i^z$, to construct a new node set $\mathcal{V}_{\mathcal{D}_t}:= \{ f\left(\bfd_{t}^i, v_i^z, \pi(t) \right) \ \vert\ i = 1, \ldots, M \}$, where $f(\cdot)$ depicts a feature assembling function, and we choose simple concatenation here. $\pi(\cdot)$ illustrates the temporal process embedding function. Then, the process sends $\mathcal{V}_{\mathcal{D}_t}$ to $U$, which comprehensively understands group dynamics according to graph edges $\mathcal{E}$, by subscribing and aggregating information from all processes. In our task, we set $U$ based on triplet-GCN.
The feature aggregation function $U(\cdot)$ is based on Eq.~\ref{eq:gcn}, with input $(\beta_{v_{i}}^{(0)}, \beta_{e_{i\to j}}^{(0)}, \beta_{v_{j}}^{(0)})=(v^{\bfd_t}_i, e_{i\to j},v^{\bfd_t}_j)$, where $v^{\bfd_t}_i$ and $v^{\bfd_t}_j$ are elements in $\mathcal{V}_{\mathcal{D}_t}$.
The denoiser of each process receives the aggregated features for conditioning. In this form, one sending and one receiving step constitute an ` information echo.' Note that the Langevin dynamics here are different from the ones in a normal diffusion model, where we introduce the relationship of a group of denoising data into the condition $\mathcal{C}_{\mathcal{D}_t}$, making it temporarily changed and steadily introducing constraints to generate next states.
With forward process variances $\beta_t$ and $\sigma_t = \sqrt{\beta_t}$ as well as the shorthand notations $\alpha_t \defeq 1-\beta_t$ and $\bar\alpha_t \defeq \prod_{s=1}^t \alpha_s$, we define one conditional step in the denoising procedure as: 
\begin{equation}
\begin{aligned}
    \bfd_{t-1}^i &= \frac{1}{\sqrt{\alpha_t}}\left( \bfd_t^i - \frac{1-\alpha_t}{\sqrt{1-\bar\alpha_t}} \varepsilon_\theta(\bfd_t^i, \pi(t), \mathcal{C}_{\mathcal{D}_t}) \right) + \sigma_t \bz, \\
    \mathcal{C}_{\mathcal{D}_t} &= U \left( \mathcal{G}_{\mathcal{D}_t} \right), \mathcal{G}_{\mathcal{D}_t} = \left\{\mathcal{V}_{\mathcal{D}_t}, \mathcal{E}\right\},
\end{aligned}
\label{echocondition}
\end{equation}
where $\bz \sim \mathcal{N}(0,1)$ if $t > 1$, else $\bz = 0$. The information echo is repeated at every timestep, collectively forming the information over the scene graph diffusion. 

\subsection{Layout Branch}
\label{lb}
As described in Sec.~\ref{echodiff}, we model the layout generation by setting each node with its own denoising process and encouraging them to interact with their diffused layout with each other at every denoising step.

\begin{figure*}[t!]
    \centering
    \includegraphics[width=0.95\linewidth]{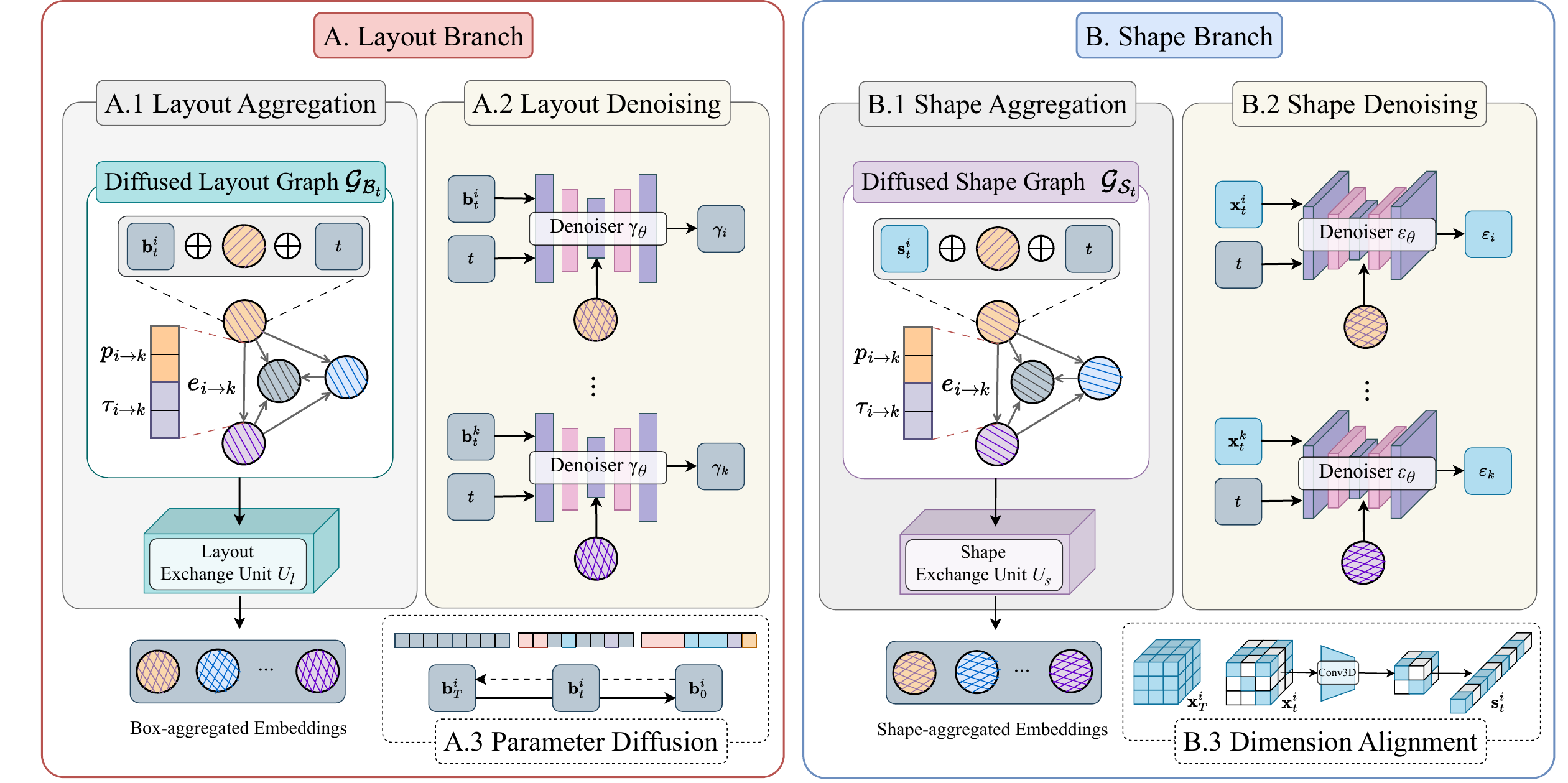}
    \caption{\textbf{One Step of Dual-Branch Information Echo.} For each time step, we encourage the layout (left) and shape (right) branches to exchange information within each branch for all objects in the same scene.
    }
    \label{fig:detail}
\end{figure*}

\myparagraph{Layout Parametrization.}
The scene layout is represented by object bounding boxes. Initially, each bounding box ${\bf b}_0^i$ has 7 parameters, e.g., location $(x,y,z)$, size $(l,h,w)$, and a yaw angle $\theta$. Before training, we normalize location and size by their maximum and minimum values on each axis. For angles, we calculate its sine and cosine values. Thereby, the final representation contains 8 parameters: ${\bf b}_0^i=\{x,y,z,l,h,w,\sin{\theta},\cos{\theta}\}$, as shown in Fig.~\ref{fig:detail}.A.3.

\myparagraph{Layout Echoes.}
Since the bounding box generation needs to be compliant with the spatial constraints described in the scene graph, state observation from other nodes is needed to determine the bounding box of a specific node. In this branch, we encourage all nodes to communicate their current diffused bounding boxes ${\bf B}_t:= \{{\bf b}_t^i~|~i=1, ..., M\}$ in the graph by a layout exchange unit $U \mapsto U_l$ to achieve global awareness at each time step. As shown in Fig.~\ref{fig:detail}.A.1, in the implementation, we substitute denoising data $\bfd_t^i$ to the diffused bounding box ${\bf b}_t^i$, resulting in  $\mathcal{V}_{\mathcal{D}_t} \mapsto \mathcal{V}_{\mathcal{B}_t}$ and $\mathcal{G}_{\mathcal{D}_t} \mapsto \mathcal{G}_{\mathcal{B}_t}$. Thus, layout echoes happen at each time step by using Eq.~\ref{echocondition}, which are essential for the functionality of this branch.

\myparagraph{Training objective.}
We follow a normal DDPM training routine, in which we set 1000 time steps for all diffusion processes with weight-shared $\gamma_\theta$. In each time step, the objective is to minimize the noise prediction errors supervised by $\gamma$ sampled from Gaussian Distribution:
\begin{equation}
    \mathcal{L}_{layout} = 
    \mathbb{E}_{{\bf B},\gamma\sim\mathcal{N}(0,1),t} 
    \left [ || \gamma - \gamma_\theta({\bf B}_t, \pi(t),  U_{l}(\mathcal{G}_{\mathcal{B}_t}) ||_2^2 \right ], \mathcal{G}_{\mathcal{B}_t} = \{\mathcal{V}_{\mathcal{B}_t}, \mathcal{E}\}.
\end{equation}
\subsection{Shape Branch}
\label{sb}
As shown in Fig.~\ref{fig:pipeline}.B.2, we pretrain a VQ-VAE as a shape en-decoder and treat the latent codes ${\bf X}_0$ in the bottleneck of the VQ-VAE as the ground truth of the LDM, as deployed in CommonScenes~\cite{commonscenes2023}. Similar to our layout branch, we introduce the information echo scheme to shape generation, enhancing the generation consistency.

\myparagraph{Isolation Problems.}
Unlike layout generation, shape generation can be conditioned solely with the semantic information~\cite{cheng2023sdfusion,li2023diffusionsdf} or relation information~\cite{commonscenes2023}, which means the shape branch is already functional when allocating each node to a denoising process driven by semantics or latent relation embeddings. However, each process still lacks consistency control from the shape appearance aspect. In other words, each generation process only focuses on its own, while ignoring others' shape generation state throughout the whole denoising. Thus, the generations are isolated from each other, resulting in suboptimal inter-object consistency of the global style.

\myparagraph{Shape Echoes.} We solve the problem by introducing shape observation of other processes for each process, which is achieved by shape echoes. As shown in Fig.~\ref{fig:detail}.B.3, before shape echoes happen, we first feed diffused shape codes ${\bf X}_t:= \{{\bf x}_t^i, ~|~i=1,..., M\}$ to 3D convolutional layers and flatten to ${\bf S}_t:= \{{\bf s}_t^i, ~|~i=1,..., M\}$, aligning the dimension with the node's attribute. The meaning of the ${\bf X}_t$ and ${\bf S}_t$ becomes more obvious when the time step is closer to 0. Similar to the layout branch, we substitute denoising data $\bfd_t^i$ to the diffused shape code ${\bf s}_t^i$, resulting in  $\mathcal{V}_{\mathcal{D}_t} \mapsto \mathcal{V}_{\mathcal{S}_t}$ and $\mathcal{G}_{\mathcal{D}_t} \mapsto \mathcal{G}_{\mathcal{S}_t}$. Thus, with a shape exchange unit $U \mapsto U_s$, shape echoes happen at each time step by using Eq.~\ref{echocondition}.

\myparagraph{Training objective.} For each latent diffusion process, the weight-shared denoiser $\varepsilon_\theta$ takes time step $t$, shape codes ${\bf X}_t$, and shape-aggregated feature as input to predict the noise, which is supervised by $\varepsilon$ sampled from Gaussian Distribution. The objective of the training is to minimize the noise prediction errors:
\begin{equation}
    \mathcal{L}_{shape} = 
    \mathbb{E}_{{\bf X},\varepsilon\sim\mathcal{N}(0,1),t} 
    \left [ || \varepsilon - \varepsilon_\theta({\bf X}_t, \pi(t),  U_{s}(\mathcal{G}_{\mathcal{S}_t})||_2^2 \right ], \mathcal{G}_{\mathcal{S}_t} = \{\mathcal{V}_{\mathcal{S}_t}, \mathcal{E}\}
\end{equation}
\subsection{Dual-Branch Joint Training}
Since both branches are based on diffusion models, the framework supports end-to-end synchronized training optimized by two losses weighted by ${\lambda_1, \lambda_2}$:
\begin{equation}
    \mathcal{L} = \lambda_1 \mathcal{L}_{layout} + \lambda_2 \mathcal{L}_{shape}.
\end{equation}
\section{Experiments}
\label{sec:exp}

\myparagraph{Dataset.} We conduct our experiments on SG-FRONT dataset~\cite{commonscenes2023}, which provides scene-graph annotations for the high-quality 3D-FRONT~\cite{3dfront} with household environments. SG-FRONT contains 15 relationship types and 45K object instances from three types of scenes.  

\myparagraph{Metrics.} We follow the metrics used in ~\cite{commonscenes2023,zhai2023sg,lin2024instructscene}, to measure the \emph{fidelity and diversity} of generated scenes, where we adopt Fréchet Inception Distance (FID)~\cite{fid2017}, $\text{FID}_{\text{CLIP}}$~\cite{kynkaanniemi2022role} and Kernel Inception Distance (KID)~\cite{kid2018} metrics \cite{Paschalidou2021NEURIPS}. To measure the \emph{scene graph consistency}, we follow the scene graph constraints~\cite{graph2scene2021}, which measure the accuracy of a set of relations on a generated layout. To measure the \emph{shape consistency and diversity} following \cite{commonscenes2023}, we test dining rooms, a typical scenario in which tables and chairs should be in suits. 

\myparagraph{Baselines.}
For retrieval-based methods \ie, a layout generation network \emph{3D-SLN}~\cite{Luo_2020_CVPR}, a progressive method to add objects one-by-one designed in~\cite{graph2scene2021}, \emph{Graph-to-Box} from~\cite{graph2scene2021}, \emph{DiffuScene}~\cite{tang2023diffuscene} with text conditioning, and \emph{InstructScene} \cite{lin2024instructscene} with graph conditioning.
For generative methods, \emph{Graph-to-3D}\cite{graph2scene2021}, modeling shape and layouts both in VAE.  CommonScenes~\cite{commonscenes2023}, which model scene layouts through VAE and generate each object via LDM conditioned on relation embeddings. \emph{EchoLayout/CommonLayout+SDFusion} that we design stacking the layout branch and a text-to-shape generation model SDFusion~\cite{cheng2023sdfusion} sequentially. More details are in Supplementary Material.

\myparagraph{Implementation Details.} The training, evaluation, and visualization are carried out on a single NVIDIA A40 GPU with 40 GB of memory. Training is optimized with AdamW with an initial learning rate of 1e-4. The values of ${\lambda_1, \lambda_2}$ are set to 1.0. More details are provided in Supplementary Material.

\subsection{Scene Fidelity}
\begin{table*}[t!]

\centering
    \scalebox{0.7}{
    \begin{tabular}{l  c  c c c | c c c | c c c }
     \toprule 
        \multirow{2}{*}{Method} & Shape & \multicolumn{3}{c}{Bedroom} & \multicolumn{3}{c}{Living room} & \multicolumn{3}{c}{Dining room}
        \\ & Representation & FID & $\text{FID}_{\text{CLIP}}$ & KID & FID & $\text{FID}_{\text{CLIP}}$ & KID & FID & $\text{FID}_{\text{CLIP}}$ & KID \\
    \midrule 
        3D-SLN~\cite{Luo_2020_CVPR} & \multirow{7}{*}{\rotatebox{90}{Retrieval}} & 57.90 & \phantom{0}5.45 & \phantom{0}3.85 & 77.82 & \phantom{0}7.02 & \phantom{0}3.65 & 69.13 & \phantom{0}7.99 & \phantom{0}6.23\\
        Progressive~\cite{graph2scene2021} &  & 58.01 & \phantom{0}5.67 & \phantom{0}7.36 & 79.84 & \phantom{0}7.41 & \phantom{0}4.24 & 71.35 & \phantom{0}8.28 & \phantom{0}6.21 \\
        Graph-to-Box~\cite{graph2scene2021} &  & 54.61 & \phantom{0}5.26 & \phantom{0}2.93 &  78.53 & \phantom{0}6.88 & \phantom{0}3.32 & 67.80 & \phantom{0}7.75 & \phantom{0}6.30 \\
        CommonLayout \cite{commonscenes2023} &  & 52.69 & \phantom{0}5.22 & \phantom{0}2.82 & 76.52 & \phantom{0}6.58 & \phantom{0}2.08 & 65.10 & \phantom{0}7.55 & \phantom{0}6.11 \\
        DiffuScene \cite{tang2023diffuscene} & & 52.02 & \phantom{0}5.01 & \phantom{0}2.52 & 81.61 & \phantom{0}7.52 & \phantom{0}\best{1.23} & 65.90 & \phantom{0}7.39 & \phantom{0}\best{0.09} \\
        InstructScene$^*$ \cite{lin2024instructscene}& & \best{45.40} & \phantom{0}\best{3.87} & \phantom{0}\second1.06 & \second75.83 & \phantom{0}\second6.98 & \phantom{0}4.15 & \second61.56 & \phantom{0}\second6.49 & \phantom{0}4.90\\ 
        \textbf{EchoLayout (Ours)} & & \second46.53 & \phantom{0}\second4.24 & \phantom{0}\best{0.33} &  \best{75.54} & \phantom{0}\best{6.35} & \second1.60 &  \best{59.66} & \phantom{0}\best{6.24} & \phantom{0}\second2.63\\
    \midrule 
    Graph-to-3D~\cite{graph2scene2021} & DeepSDF~\cite{park2019deepsdf} & 63.72 & \phantom{0}6.01 & 17.02 & 82.96 & \phantom{0}7.80 & 11.07 & 72.51 & \phantom{0}7.25 & 12.74\\
    CommonLayout+SDFusion~\cite{cheng2023sdfusion}& txt2shape & 68.08 & \phantom{0}5.61 & 18.64 & 85.38 & \phantom{0}7.23 & 10.04 & 64.02 & \phantom{0}6.92 & \phantom{0}5.08\\
    EchoLayout+SDFusion~\cite{cheng2023sdfusion} & txt2shape & 57.68 & \phantom{0}4.96 & 10.54 & 83.66 & \phantom{0}\second6.83 & \phantom{0}9.62 & \second65.55 & \phantom{0}\second7.02 & \phantom{0}\second4.99\\
    CommonScenes~\cite{commonscenes2023} & rel2shape & \second57.68 & \phantom{0}\second4.86 & \phantom{0}\second6.59 & \second80.99 & \phantom{0}7.05 & \phantom{0}\second6.39 & 65.71 & \phantom{0}7.04 & \phantom{0}5.47\\
    \textbf{EchoScene (Ours)} & echo2shape &  \best{48.85} & \phantom{0}\best{4.26} & \phantom{0}\best{1.77} &  \best{75.95} & \phantom{0}\best{6.73} & \phantom{0}\best{0.60} &  \best{62.85} & \phantom{0}\best{6.28} & \phantom{0}\best{1.72}\\
    \bottomrule
    \end{tabular}
    }
\caption{\textbf{Scene generation realism} as measured by FID, $\text{FID}_{\text{CLIP}}$ and KID $(\times 0.001)$ scores at $256^{2}$ pixels between the top-down rendering of generated and real scenes, following~\cite{tang2023diffuscene,commonscenes2023} (lower is better). We color the \colorbox{orange}{best} and the \colorbox{light_yellow}{second}. Two main rows are separated with respect to the reliance on an external shape database. \emph{InstructScene$^*$} represents its 3D layout decoder solely. Both \emph{EchoLayout} and \emph{CommonLayout} refer to the single layout branch.}
\label{tab:fidkid}
\end{table*}
As a generation task, the fidelity of scene synthesis is essential to measure.

\myparagraph{Quantitative results.} We perform two types of scene synthesis, which are shape retrieval-based and generation-based synthesis, and provide FID / $\text{FID}_{\text{CLIP}}$ / KID in~\cref{tab:fidkid}. For retrieval, our layout branch $EchoLayout$ performs better than the layout branch of the previous SoTA CommonLayout~\cite{commonscenes2023} across all metrics, and concurrent work DiffuScene~\cite{tang2023diffuscene} on most of the metrics. In terms of shape generation, our 3D scene generation version $EchoScene$ shows clear advantages against CommonScenes~\cite{commonscenes2023}, for example, improving FID by $15\%$, $\text{FID}_{\text{CLIP}}$ by $12\%$ and KID by $73\%$ in the bedroom generation. Notably, although retrieval-based methods overall score better due to the mesh alignment with the test database, EchoScene remains competitive, showcasing generative effectiveness. Moreover, our text-to-shape baseline \emph{Echolayout+SDFusion}  outperforms the equivalent CommonLayout+SDFusion. Despite under the issue of shape inconsistency, it still achieves similar results as CommonScenes. This also underscores the effectiveness of our layout branch, driven by layout echoes, in generating more coherent scene layouts.
For shape consistency, we test the Chamfer Distance (CD) between two generated objects that are supposed to be identical (\ie shape consistency \cite{commonscenes2023}). We report common objects in three room types in~\cref{tab:objs}, showing that our shape echoes alleviate the isolation problem in CommonScenes and thus bring better inter-object consistency.
\begin{figure*}[t!]
    \centering
    \includegraphics[width=0.8\linewidth]{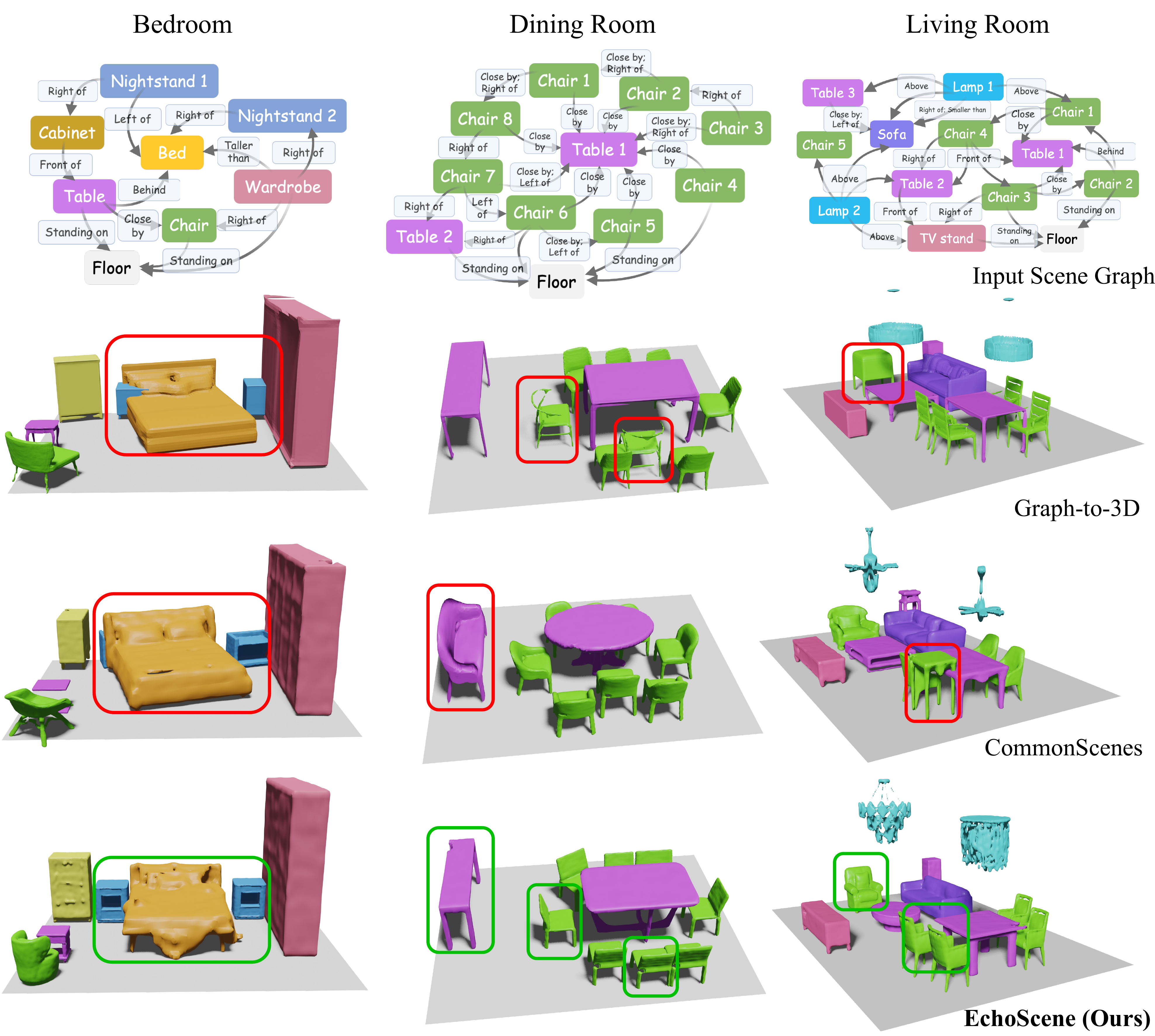}
    \caption{\textbf{Comparisons with other generative methods.} Input scene graphs have more edges between two nodes than the ones visualized here. Red rectangles highlight the inconsistent generation. (Zoom for details)
    }
    \label{fig:qualitative}
\end{figure*}

\myparagraph{Qualitative results.}  Graph-to-3D~\cite{graph2scene2021} and CommonScenes~\cite{commonscenes2023} have the same settings as EchoScene, which belong to (semi-/full-)generative models conditioned on scene graphs for the whole scene generation. We provide a comparison of these methods for each different room type in Fig.~\ref{fig:qualitative}. For example, in the bedroom, beds and nightstands of Graph-to-3D and CommonScenes are in twisted poses, indicating VAE-based layout generator does not learn angles well, while our diffusion-based architecture can provide neat results. In the dining room, Graph-to-3D generates chairs in different styles, which breaks the inter-object consistency. Although CommonScenes can generate chairs in a suit, the generated table on the left is problematic, having a chair-like shape, which makes an unsatisfactory global appearance as well. Ours achieves higher shape consistency. 
\subsection{Graph Constraints}
In this part, we evaluate the layout generation performance with respect to scene graph constraints. We compare against the closest methods: 3D-SLN~\cite{Luo_2020_CVPR}, Graph-to-3D~\cite{graph2scene2021}, InstructScene~\cite{lin2024instructscene}, and CommonScenes~\cite{commonscenes2023}. 

We optionally manipulate scene graphs in the latent by either adding a node and relevant edges to the graph or changing relations between two nodes, as illustrated in Fig.~\ref{fig:pipeline}. As shown in the top and middle row of~\cref{tab:sgconst}, for both node addition and relation change, EchoLayout and EchoScene perform better on most metrics. 3D-SLN, CommonScenes, and Graph-to-3D, for instance, loose the constraints \texttt{smaller/bigger than} and \texttt{close by} after graph manipulation, while ours largely keeps it. We believe that the improvement is caused by our scene graph diffusion strategy, where a graph-consistent sample is generated within a realistic appearance distribution. 
\begin{table*}[h!]
    \centering
    \scalebox{0.7}{
    \begin{tabular}{l c |c| cccc cc}
    \toprule
    \multirow{2}{*}{Method} & \multirow{2}{*}{\makecell{Shape \\ Representation}} & \multirow{2}{*}{\makecell{Mode}} & \multirow{2}{*}{\makecell{left/\\right}} & \multirow{2}{*}{\makecell{front/\\behind}} & \multirow{2}{*}{\makecell{smaller/\\larger}} & \multirow{2}{*}{\makecell{taller/\\shorter}} & \multirow{2}{*}{\makecell{close by}} & \multirow{2}{*}{\makecell{symmetrical}} \\
    & & & & & & & &\\
    \midrule 
        3D-SLN~\cite{Luo_2020_CVPR} & \multirow{5}{*}{\rotatebox{90}{Retrieval}} & \multirow{9}{*}{\rotatebox{90}{Change}} & 0.89 & 0.90 & 0.55 & 0.58 & 0.10 & 0.09  \\
        Progressive~\cite{graph2scene2021} & && 0.89 & 0.89 & 0.52 & 0.55 & 0.08 & 0.09  \\
        Graph-to-Box~\cite{graph2scene2021} & & & 0.91 & 0.91 & 0.86 & 0.91 & 0.66 & 0.53  \\
        CommonLayout~\cite{commonscenes2023} & & & 0.91 & 0.92 & 0.86 & 0.92 & 0.70 & 0.53  \\
        \textbf{EchoLayout (Ours)} & & & \best{0.94} & \best{0.93} & \best{0.92} & \best{0.92} & \best{0.72} & \best{0.56}\\
        \cmidrule{1-2} \cmidrule{4-9}
        Graph-to-3D~\cite{graph2scene2021} &  DeepSDF~\cite{park2019deepsdf} & & 0.91 & 0.92 & 0.86 & 0.89 & 0.69 & 0.46 \\
        CommonScenes~\cite{commonscenes2023} & rel2shape & & 0.91 & 0.92 & 0.86 & 0.91 & 0.69 & \best{0.59} \\
         \textbf{EchoScene (Ours)} & echo2shape & & \best{0.94} & \best{0.96} & \best{0.92} & \best{0.93} & \best{0.74} & 0.50\\
        \midrule \midrule
        3D-SLN~ \cite{Luo_2020_CVPR} & \multirow{5}{*}{\rotatebox{90}{Retrieval}} & \multirow{9}{*}{\rotatebox{90}{Addition}} & 0.92 & 0.92 & 0.56 & 0.58 & 0.05 & 0.05 \\
        Progressive~\cite{graph2scene2021} & &  & 0.92 & 0.91 & 0.53 &  0.54 & 0.02 & 0.06 \\
        Graph-to-Box~\cite{graph2scene2021} &  & & 0.94 & 0.93 & 0.90 & 0.94 & 0.67 & 0.58 \\
        CommonLayout~\cite{commonscenes2023} & & & 0.95 & 0.95 & 0.90 & 0.94 & 0.73 & \best{0.63}\\
        \textbf{EchoLayout (Ours)} & & & \best{0.98} & \best{0.99} & \best{0.97} & \best{0.96} & \best{0.74} & 0.58\\
        \cmidrule{1-2} \cmidrule{4-9}
        Graph-to-3D~\cite{graph2scene2021} &  DeepSDF~\cite{park2019deepsdf} & &  0.94 & 0.95 & 0.91 & 0.93 & 0.63 & 0.47\\
        CommonScenes~\cite{commonscenes2023} & rel2shape & & 0.95 & 0.95 & 0.91 & 0.95 & 0.70 & \best{0.61}\\
         \textbf{EchoScene (Ours)} & echo2shape & & \best{0.98} & \best{0.99} & \best{0.96} & \best{0.96} & \best{0.76} & 0.49\\
    \midrule
    \midrule 
        3D-SLN~\cite{Luo_2020_CVPR} & \multirow{6}{*}{\rotatebox{90}{Retrieval}} & \multirow{10}{*}{\rotatebox{90}{None}} & 0.97 & 0.99 & 0.95 & 0.91 & 0.72 & 0.47 \\
        Progressive~\cite{graph2scene2021} &  & & 0.97 & 0.99 & 0.95 & 0.82 & 0.69 & 0.46 \\ 
        Graph-to-Box~\cite{graph2scene2021} &  & & 0.98 & 0.99 & 0.96 & 0.95 & 0.72 & 0.45 \\
        CommonLayout \cite{commonscenes2023} & & & 0.98 & 0.99 & \best{0.97} & 0.95 & \best{0.74} & 0.63 \\
        InstructScene$^*$ \cite{lin2024instructscene} & & & 0.99 & 0.99 & 0.85 & 0.86 & 0.52 & 0.58 \\
        \textbf{EchoLayout (Ours)} & & & \best{1.00} & 0.99 & 0.95 & \best{0.96} & \best{0.74} & \best{0.67} \\
    \cmidrule{1-2} \cmidrule{4-9}
        Graph-to-3D~\cite{graph2scene2021} & DeepSDF~\cite{park2019deepsdf} & & 0.98 & 0.99 & 0.97 & 0.95 & 0.74 & 0.57\\
        CommonScenes \cite{commonscenes2023} & rel2shape & & 0.98 & \best{1.00} & 0.97 & 0.95 & \best{0.77} & \best{0.60} \\
        \textbf{EchoScene (Ours)} & echo2shape & & 0.98 & 0.99 & 0.96 & \best{0.96} & 0.74 & 0.55\\
    \bottomrule
    \end{tabular}}
    \caption{\textbf{Scene graph constraints} (higher is better). Top: Relationship change mode. Middle: Node addition mode. Bottom: No manipulation (\ie, generation only).}
    \label{tab:sgconst}
\end{table*}

The decrease in \texttt{symmertical} category compared with CommonScenes is likely caused by the same effect. Inspection of the data reveals that \texttt{symmertical} is an extremely rare label in the supplementary of CommonScenes~\cite{commonscenes2023}, with an occurrence of only 0.9\%. Thus, the learned part of the distribution is underrepresented and less accurate. The residual echo correction diffuses the latent code towards the learned distribution, resulting in a slight variation. In the bottom row, we report the results without manipulation. All baselines show good performance, yet our EchoLayout and EchoScene are still competitive on their own track. Notably, even though InstructScene shows impressive performance on the global generation fidelity shown in~\cref{tab:fidkid}, it struggles to maintain the multiple graph constraints simultaneously, as we described in~\cref{sec:intro}, while our information echo scheme can sufficiently aggregate relations and condition each process to achieve better results.
Distribution-dependant effects from data equally manifest in the results where we can see some drop for \texttt{symmertical} again. Interestingly, the effect is less pronounced for a lower parametric final task such as layout generation. This can be explained by the fact that the variability of bounding boxes is smaller compared to shape variations.

\subsection{Ablation study}
We ablate three parts and report the performance in~\cref{tab:abl}. Firstly, we ablate $\pi(t)$ to check the influence on the model without explicit time information. We observe a marginal decrease, indicating even without $\pi(t)$, the model still learns temporal information through every denoising step. Secondly, we exclude the shape echoes to degrade the shape branch. We clearly see a drop in the fidelity evaluation, where `mSG' focusing on the layouts is not influenced. Thirdly, we substitute cross-attention for concatenation to check if the condition way influences the information echo scheme. We also observe a drop in all metrics. 

\begin{figure*}[t!]
    \centering
    \includegraphics[width=0.7\linewidth]{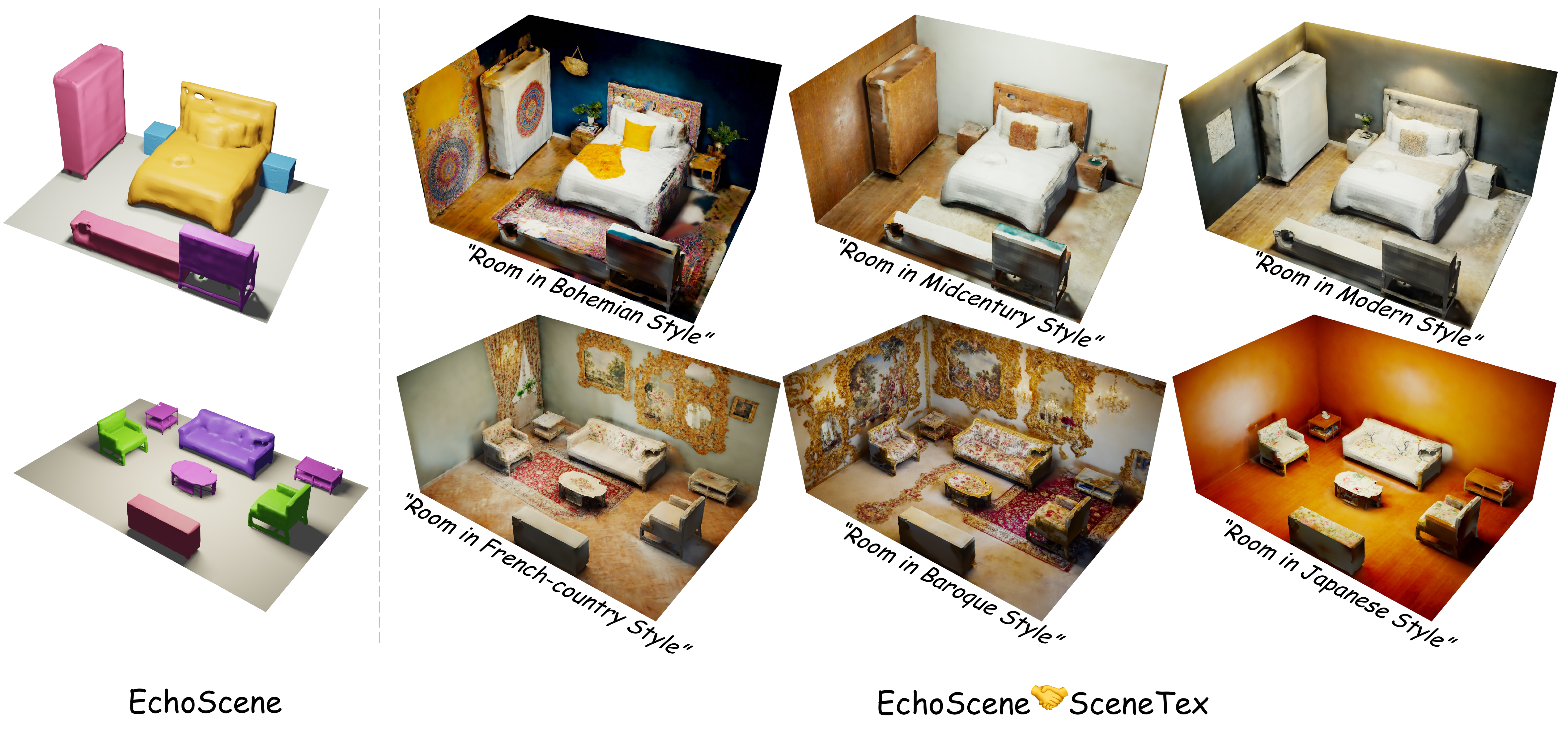}
    \caption{\textbf{Off-the-shelf texture creation.} A bedroom (top) and living room (bottom) generated by EchoScene and textured in different styles by SceneTex~\cite{chen2023scenetex}. 
    }
    \label{fig:scenetex}
\end{figure*}
\begin{figure}[t!]
    \centering
    \begin{minipage}[t]{0.5\textwidth}
        \centering
        \scalebox{0.7}{
            \begin{tabular}{l cc  cc  cc}
                \toprule
                     \multirowcell{2}{Method} & \multicolumn{2}{c}{Bedroom} & \multicolumn{2}{c}{Living room} & \multicolumn{2}{c}{Dining room}\\
                     \cmidrule{2-7}
                     & Ward. & N.stand & Chair & Table & Chair & Table\\
                 \midrule
                     CommonScenes \cite{commonscenes2023} & 0.61 & 2.69 & 6.64 & 11.75 & 1.96  & 9.04 \\
                     \textbf{EchoScene} & \best{0.14} & \best{1.68} & \best{0.99} & \best{3.02} & \best{1.75}  & \best{1.26}\\
                 \bottomrule
            \end{tabular}
        }
        \captionof{table}{\textbf{Inter-object Consistency.} The consistent object shapes within a scene are indicated by low CD values ($\times 0.001$).\label{tab:objs}}
    \end{minipage}%
    \hspace{2mm}
    \begin{minipage}[t]{0.45\textwidth}
        \centering
        \scalebox{0.65}{
            \begin{tabular}{lcccc}
                \toprule
                     Ablation & FID & $\text{FID}_{\text{CLIP}}$ & KID & mSG \\
                 \midrule
                     Ours w/o $\pi(t)$ & 40.55 & \best3.14 & 1.69 & 0.87\\
                     Ours w/o shape echoes & 46.88 & 3.81 & 4.17 & \best0.88\\
                     Ours with concat & 48.32 & 3.82 & 6.87 & 0.87\\
                     \textbf{Ours} & \best39.74 & \best3.14& \best1.24 & \best0.88\\
                 \bottomrule
            \end{tabular}
        }
        
        \captionof{table}{Ablations under three circumstances. mSG means average graph constraints.\label{tab:abl}}
    \end{minipage}
\end{figure}

\subsection{Application}
The limitation of EchoScene is that it can generate semantic yet only textureless scenes, which prohibits it from downstream tasks requiring photorealistic textures. Yet, the limitation can be alleviated by directly equipping EchoScene with a subsequent off-the-shelf texture generator. Here, we demonstrate that textured scenes can be generated via SceneTex~\cite{chen2023scenetex} in~\cref{fig:scenetex}.

\section{Conclusion} 
We have introduced EchoScene, a scene-generative method based on dual-branch diffusion models, which are layout branch and shape branch. To make two branches functional, we have proposed an information scheme encouraging information exchange within both branches at each denoising step, generating final scenes that are compliant with the scene graph description. Experiments show that EchoScene can achieve higher generation fidelity and more inter-object consistency of the global scene style.


\section*{Acknowledgements}
We are grateful to Google University Relationship GCP Credit Program for supporting this work by providing computational resources.
%
%
\bibliographystyle{splncs04}
\bibliography{main}

\appendix
\section*{Supplementary Material of EchoScene}
In this Supplementary Material, we report the following:
\begin{itemize}
\item Section~\ref{sec:mani}: Manipulation Strategy.
\item Section~\ref{sec:manipulation}: Manipulation Qualitatives.
\item Section~\ref{sec:results}: Additional Results.
\item Section~\ref{sec:qual}: Additional Qualitatives.
\item Section~\ref{sec:training}: Additional Experimental Details.
\end{itemize}
\section{Manipulation Strategy}
\label{sec:mani}
We propose a novel training strategy for node and edge manipulation tailored for scene graph diffusion, thus editing the generated scenes.

\begin{figure*}[t]%
    \centering
    \includegraphics[width=0.95\linewidth]{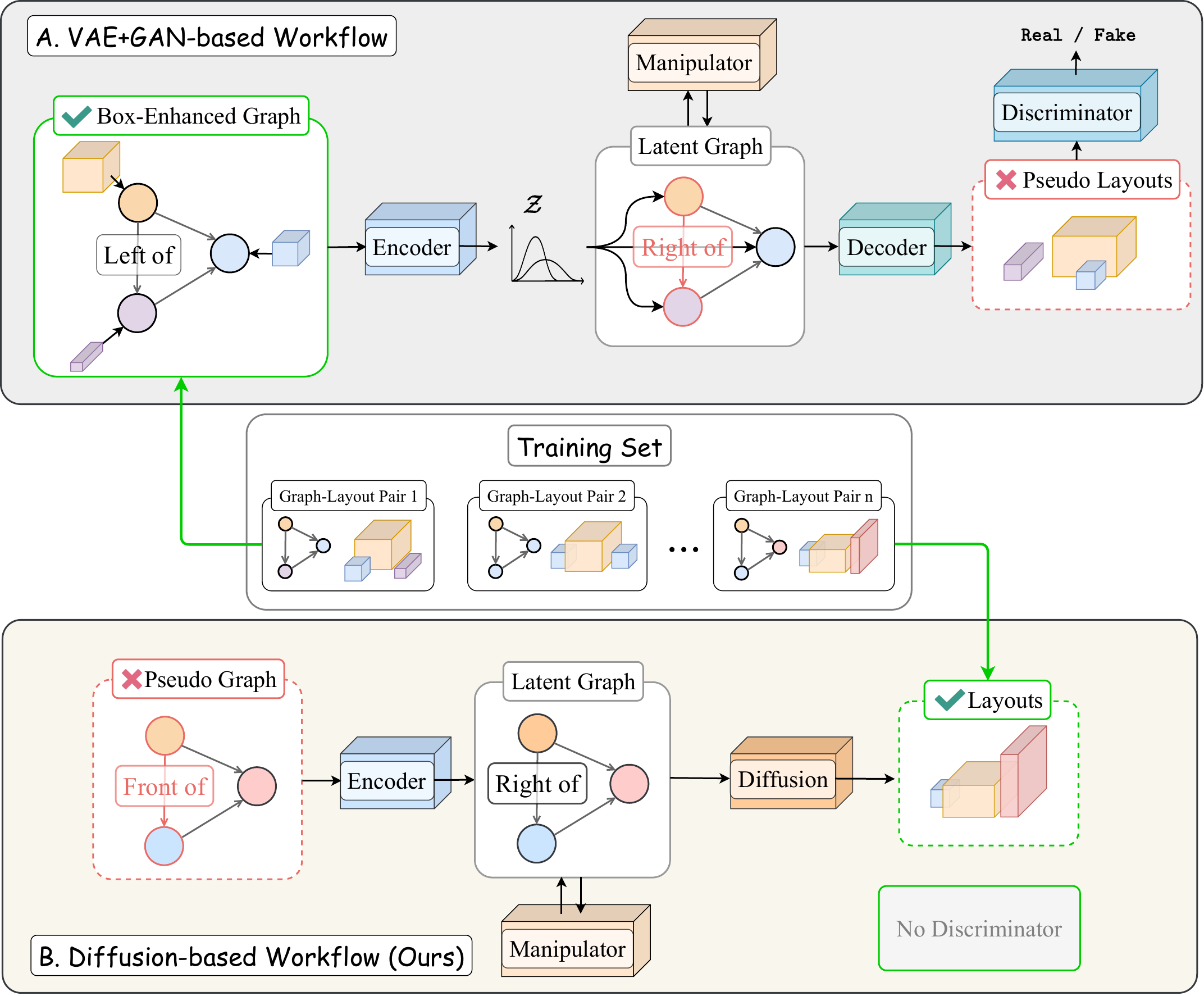}
    \caption{\textbf{Manipulation workflow.} (Top) Previous VAE-based methods set the input as real and use a GAN-based module to facilitate the output prediction. (Bottom) Our Diffusion-based framework inverse the workflow by starting with a pseudo graph and ending with ground truth, without redundant modules.}
    \label{fig:mani}
    \vspace{-5mm}
\end{figure*}

\myparagraph{VAE$+$GAN-based strategy.}
In the previous methods~\cite{commonscenes2023,graph2scene2021}, the layout branch is modeled by a VAE architecture. During training, the input scene graph must be augmented by ground truth bounding boxes. In this case, pseudo layouts are created from the output side after the manipulation, and there is no matching ground truth to supervise them. To make the pipeline functional, they set an additional GAN module to help train the pseudo label. For example, as shown in Fig.~\ref{fig:mani}.A, the original relationship is \{\texttt{Bed} $\longrightarrow$ \texttt{Left of} $\longrightarrow$ \texttt{TV stand} \}, with a manipulator changing it to \{\texttt{Bed} $\longrightarrow$ \texttt{Right of} $\longrightarrow$ \texttt{TV stand} \}. Only the bounding boxes on the input side exist in the dataset, while the ones on the output side are generated and need to be discriminated.
Such a manipulation strategy relies on the performance of GAN, whose training scheme is not stable and requires large-scale training, consequently affecting the training quality of layout VAE. Thus, the results shown in Tab. 2 in the main paper are always unsatisfactory.

\myparagraph{Our strategy.}
In contrast, we set both branches through diffusion processes, avoiding the necessity of input ground truth bounding boxes during training. Instead, we are able to simulate an inverted workflow by setting input as pseudo graphs with fake relations, and we manipulate them back to the existing graphs in the dataset. For example, as shown in Fig.~\ref{fig:mani}.B, the ground truth relation label is \{\texttt{Bed} $\longrightarrow$ \texttt{Right of} $\longrightarrow$ \texttt{Nightstand}\}, we simulate the relation changing by setting an arbitrary label, \textit{e.g.}, \{\texttt{Bed} $\longrightarrow$ \texttt{Front of} $\longrightarrow$ \texttt{Nightstand}\}. In this case, we do not need to provide real data as input and further do not set an additional GAN module to facilitate the training. Thus, the arbitrary pseudo side enables large-scale training, maintaining a good manipulation performance. More importantly, we do not need to discriminate the final layout prediction. With our proposed strategy, we maintain the typical diffusion training routine, which has Gaussian noise as supervision during training without further exhausting inference. Last but not least, pure diffusion dynamics guarantee stable training. 

\begin{figure*}[!t]
    \centering
    \includegraphics[width=\linewidth]{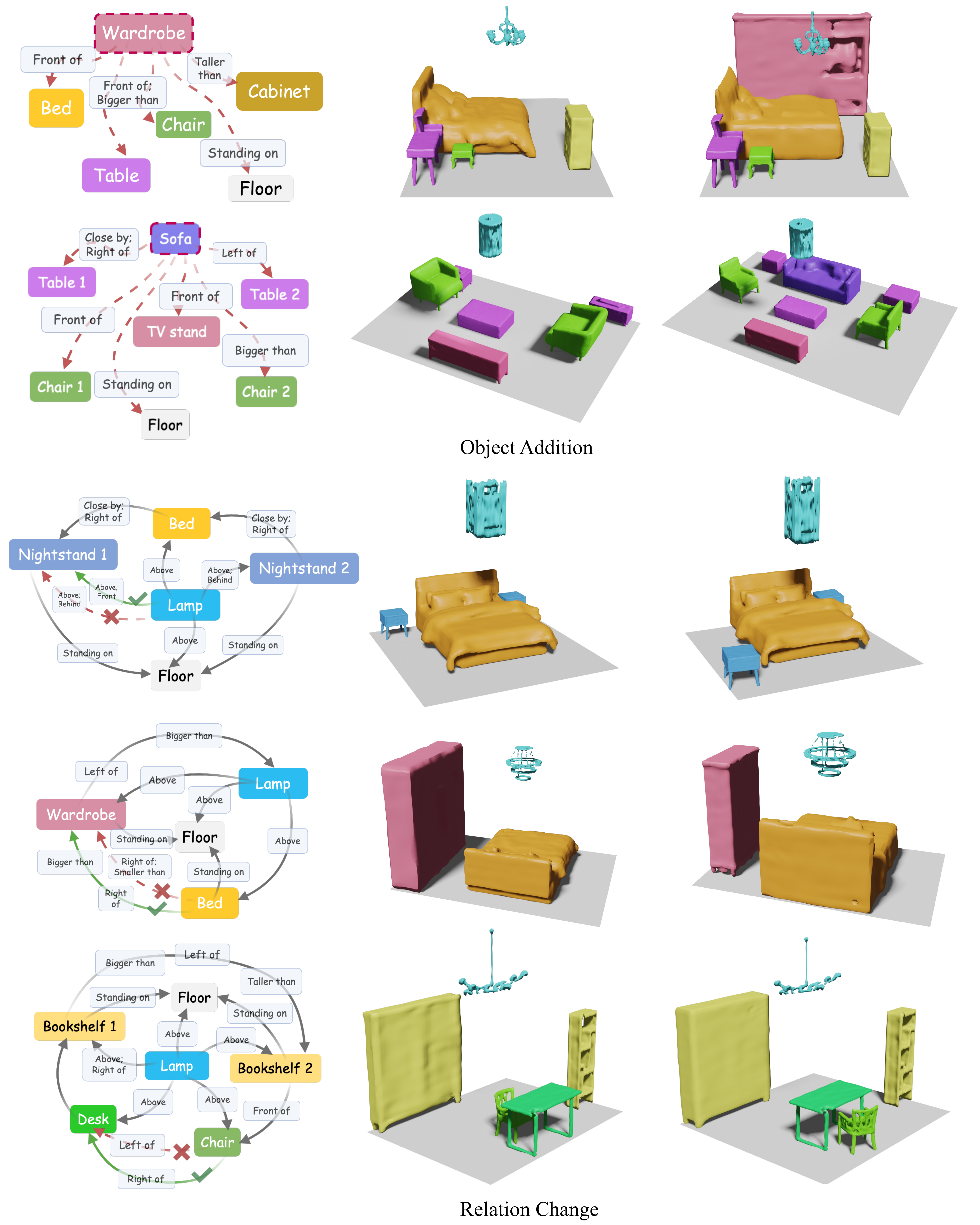}
    \caption{\textbf{Scene manipulation.} (Upper main row) Object addition to the scene graph. (Bottom main row) Partial relation change in the graph. (Zoom for details)
    }
    \label{fig:mani_qua}
    \vspace{-5mm}
\end{figure*}
\section{Manipulation Qualitatives}
\label{sec:manipulation}
We provide several qualitative results of the scenes before and after manipulation in Fig.~\ref{fig:mani_qua}. The procedure includes EchoScene generating a scene based on a provided scene graph, followed by scene modification according to adjustments made to the graph. 

\myparagraph{Object addition.}
We first show object addition in the scene through the incorporation of new nodes and their corresponding edges into the scene graph. For instance, the second row illustrates the addition of a sofa node to the graph. Upon this addition, EchoScene adeptly reconfigures the scene, modifying the pose and appearance of pre-existing objects to seamlessly integrate the sofa into the scene. 

\myparagraph{Relation change.}
Next, we show object rearrangement within the scene through modifications to specific graph relations. The depiction is organized row-wise, showcasing various types of edge manipulations: the first row highlights the \texttt{front of/behind} relationship, the second focuses on \texttt{bigger/smaller than}, and the final row on \texttt{left/right of}. 
An interesting case is observed in the second row, where the bed is adjusted to be bigger than the wardrobe from the volume perspective. This particular relationship between the bed and wardrobe is comparatively rare within the dataset. 
Yet, EchoScene can jointly adjust the bounding box sizes of both the wardrobe and bed to achieve the goal. In the last row, EchoScene successfully alters the relationship between the desk and chair, while reorienting the chair's pose to face the desk, thereby maintaining inter-object consistency within the scene.

\section{Additional Results}
\label{sec:results}
We believe our shape branch driven by shape echoes can bring more object generation compliance to the global scenes. Thus, we further conduct object-level analysis, following~\cite{commonscenes2023}, to report the MMD ($\times0.01$), COV (\%), and 1-nearest neighbor accuracy (1-NNA, \%) for evaluating per-object generation. As shown in the first two rows of Table~\ref{tab:a}, our method shows better performance on most of the categories in both MMD and COV, which highlights the object-level shape generation ability of EchoScene. 
1-NNA directly measures distributional similarity to the ground truth objects in both diversity and quality. The closer 1-NNA is to 50\%, the better the shape distribution is captured. It can be observed that in most of the categories, our method surpasses CommonScenes in the evaluation of distributional similarity. Overall, EchoScene exhibits more plausible object-level generation than the previous state-of-the-art. 
\begin{table*}[t]
    \centering
    \scalebox{0.8}{
    \begin{tabular}{l c|cccccccccc}
    \toprule
    Method & Metric & Bed & N.stand & Ward. & Chair & Table & Cabinet & Lamp & Shelf & Sofa & TV stand\\
    \midrule 
        Graph-to-3D~\cite{graph2scene2021} & \multirow{3}{*}{MMD ($\downarrow$)} & 1.56 & 3.91 & 1.66 & 2.68 & 5.77 & 3.67 & 6.53 & 6.66 & 1.30 & 1.08 \\
        CommonScenes~\cite{commonscenes2023}& & {0.49} & {0.92} & {0.54} & 0.99 & {1.91} & {0.96} & {1.50} & {2.73} & {0.57} & \best{0.29} \\
        \textbf{EchoScene (Ours)} & & \best0.37 & \best0.75  & \best0.39 & \best0.62 & \best1.47 & \best0.83 & \best0.66 & \best2.52 & \best0.48 & 0.35 \\
    \midrule 
        Graph-to-3D~\cite{graph2scene2021} & \multirow{3}{*}{COV ($\%, \uparrow$)} & 4.32 & 1.42 & 5.04 & 6.90 & 6.03 & 3.45 & 2.59 & 13.33 & 0.86 & 1.86 \\
        CommonScenes~\cite{commonscenes2023}& & {24.07} & {24.17} & {26.62} & \best{26.72} & \best{40.52} & {28.45} & \best{36.21} & {40.00} & {28.45} & {33.62} \\
        \textbf{EchoScene (Ours)} & & \best39.51 & \best25.59 & \best37.07 & 17.25 & 35.05 & \best43.21 & 33.33 & \best50.00 & \best41.94 & \best40.70\\
    \midrule 
        Graph-to-3D~\cite{graph2scene2021} & \multirow{3}{*}{1-NNA ($\%, \downarrow)$} & 98.15 & 99.76 & 98.20 & 97.84 & 98.28 & 98.71 & 99.14 & 93.33 & 99.14 & 99.57 \\
        CommonScenes~\cite{commonscenes2023}& & {85.49} & {95.26} & {88.13} & \best{86.21} & \best{75.00} & {80.17} & \best{71.55} & {66.67} & {85.34} & {78.88} \\
        \textbf{EchoScene (Ours)} & & \best72.84 & \best91.00 & \best81.90 & 92.67 & 75.74 & \best69.14 & 78.90 & \best35.00 & \best69.35 & \best78.49\\
    \bottomrule
    \end{tabular}
    }
    \caption{\textbf{Object-level generation performance.} We report MMD($\times0.01$), COV and 1-NNA for evaluating shapes by means of quality and diversity.}
    \label{tab:a}
\end{table*}
\begin{figure*}[t!]
    \centering
    \includegraphics[width=\linewidth]{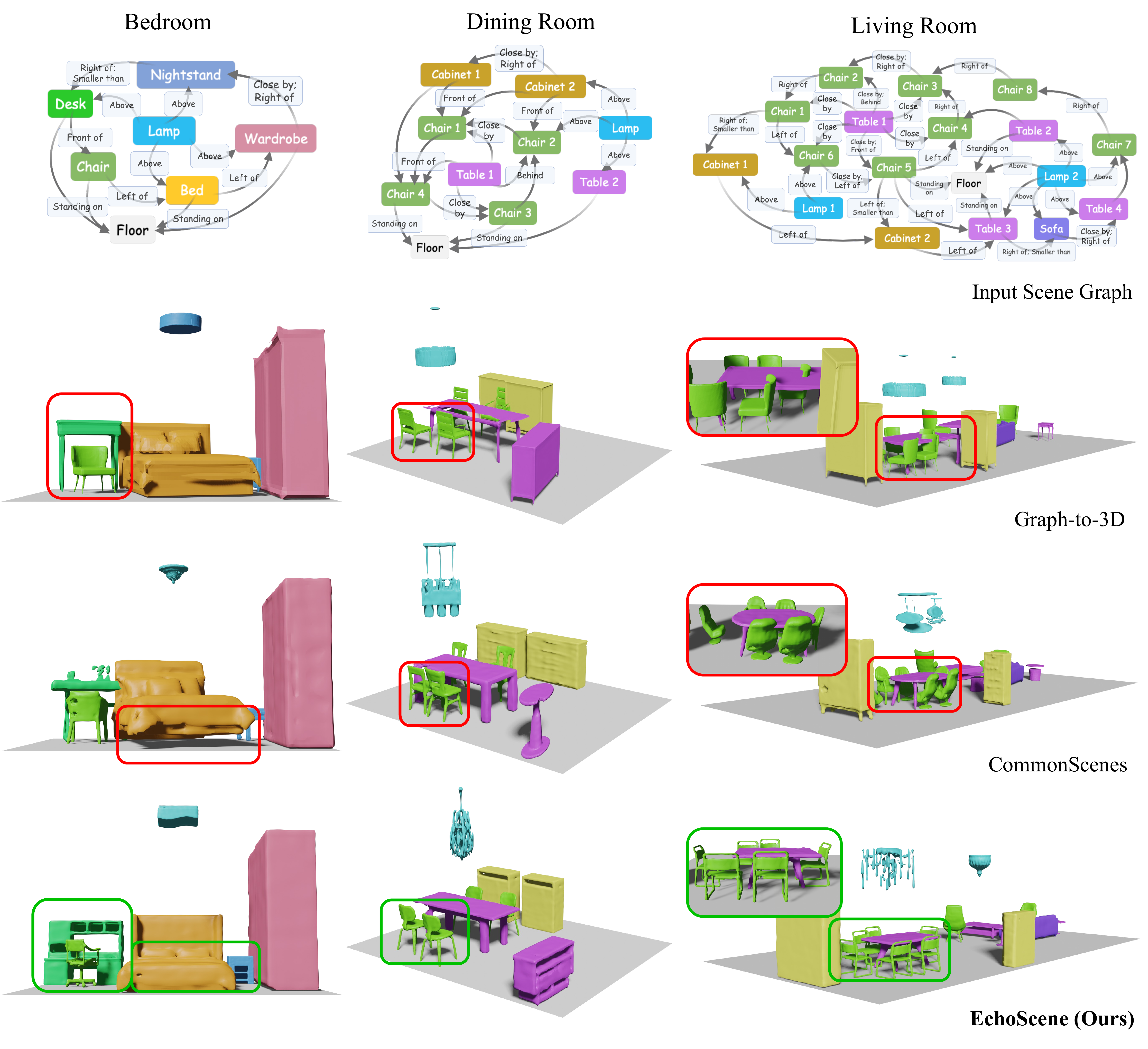}
    \caption{\textbf{More comparisons with other methods.} Red rectangles highlight the inconsistent generation. (Zoom for details)
    }
    \label{fig:addition}
    \vspace{-5mm}
\end{figure*}
\begin{figure*}[t!]
    \centering
    \includegraphics[width=0.95\linewidth]{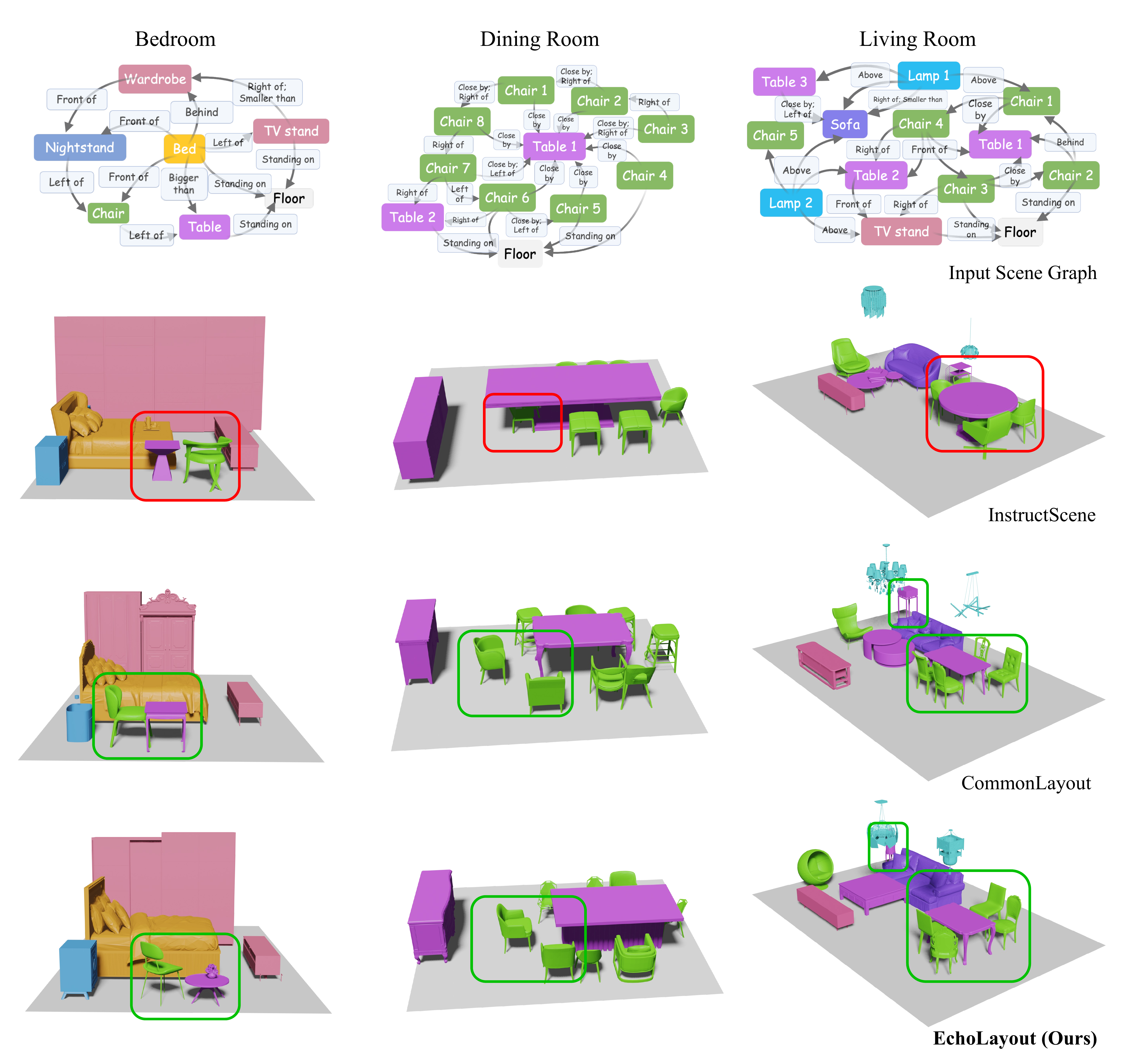}
    \caption{\textbf{Comparisons with other retrieval methods.} Input scene graphs have more edges between two nodes than the ones visualized here. Red rectangles highlight the inaccurate graph constraints. (Zoom for details)
    }
    \label{fig:retrieval}
\end{figure*}
\begin{figure*}[t!]
    \centering
    \includegraphics[width=\linewidth]{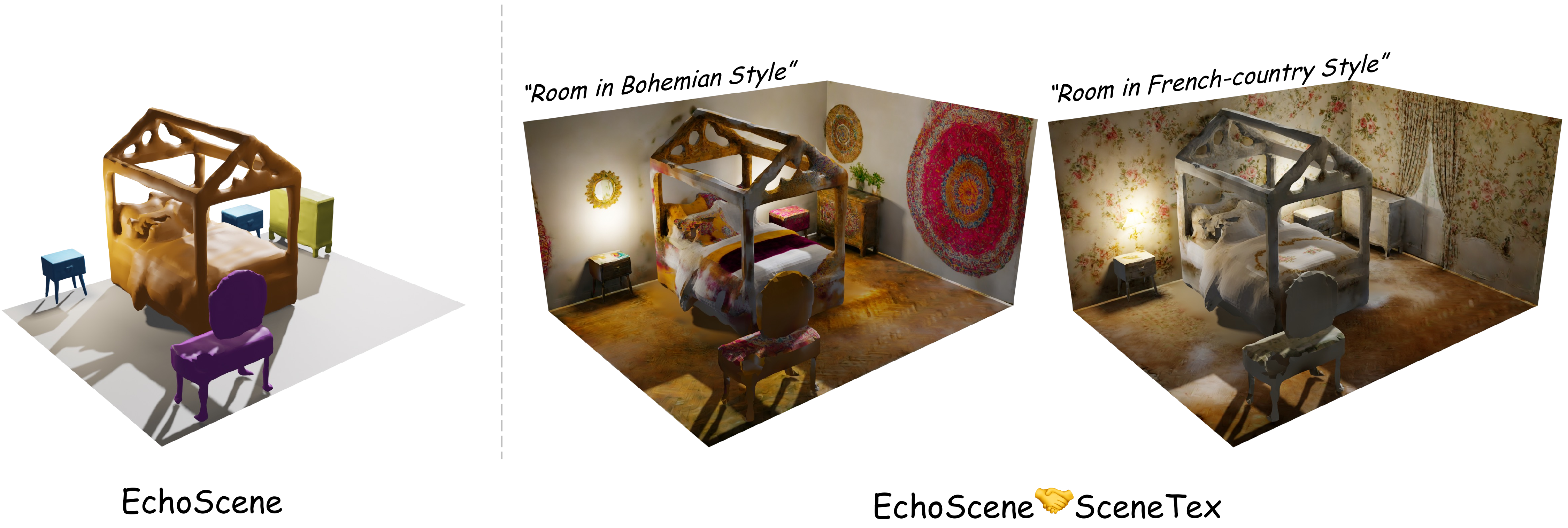}
    \caption{\textbf{Off-the-shelf texture creation.} A bedroom with a complex structure bed inside generated by EchoScene and textured in different styles by SceneTex~\cite{chen2023scenetex}. 
    }
    \label{fig:scenetex_supp}
\end{figure*}
\section{Additional Qualitatives}
\label{sec:qual}
\myparagraph{Generative methods.} We first show more quantitative results in Fig.~\ref{fig:addition}. Our method can achieve more inter-object consistency and generation quality. For example, in the bedroom, Graph-to-3D fails to achieve desk-and-chair consistency, while EchoScene can generate a desk whose appearance is closer to a desk in the real scenario with a chair suitable for it. In the living and dining rooms, Graph-to-3D either fails on shape consistency or angle predictions of chairs. CommonScenes cannot guarantee a fine-grained shape consistency, while the chairs coarsely look similar. In contrast, EchoScene can generate coherent shapes and make pose prediction more accurate.

\myparagraph{Retrieval methods.} 
Retrieval methods select objects from the database based on how closely their bounding box sizes match those of the generated layouts. Therefore, this line of work suffers from inter-object inconsistency; for example, chairs are not recalled in suits. Such inconsistencies often stem from even minor misalignment in the size of generated bounding boxes, leading to the selection of entirely different objects than intended. Despite this, our focus in Figure~\ref{fig:retrieval} is to illustrate the effectiveness of graph constraints, instead of consistency. It is observable that even though InstructScene~\cite{lin2024instructscene} demonstrates the capability to generate objects in a decent manner, it fails to adhere to the partial graph constraints. On the contrary, both CommonLayout~\cite{commonscenes2023} and EchoLayout exhibit proficiency in complying with these constraints.

\myparagraph{Texture Generation.} 
We finally show additional texture generation on a relatively complex structured bedroom in Fig.~\ref{fig:scenetex_supp}. EchoScene can provide well-generated scenes that are compatible with an off-the-shelf SceneTex~\cite{chen2023scenetex} to generate textures.

\section{Additional Experimental Details}
\label{sec:training}

\subsection{Baselines.}
\myparagraph{Graph-to-3D series~\cite{graph2scene2021}.} This series includes one generative method and three object retrieval methods. First, the full generative version \emph{Graph-to-3D}, stacking two VAE-based branches for object and layout generation, respectively. Second, \emph{Graph-to-Box}, the single layout branch focusing on the object retrieval task. Third, \emph{Progressive}, a modified baseline upon Graph-to-Box, specifically adding objects one by one in an autoregressive manner. Fourth, \emph{3D-SLN}~\cite{Luo_2020_CVPR}, sharing the same architecture as Graph-to-Box, but without layout standardization during training. We follow the illustration in the supplementary of CommonScenes.

\myparagraph{CommonScenes series~\cite{commonscenes2023}.} This series includes the fully generative version \emph{CommonScenes}, and its layout branch for object retrieval, \emph{CommonLayout}. We follow the illustration in the supplementary of CommonScenes.

\myparagraph{Text-to-shape series.} This series includes two generative baselines. One is built upon CommonScenes called \emph{CommonLayout+SDFusion}, and the other is \emph{EchoLayout+SDFusion}, with EchoLayout as our layout branch. Both methods achieve the fully generative ability by first generating the bounding boxes and further generating shapes upon a text-to-shape method SDFusion~\cite{cheng2023sdfusion} within each bounding box, according to the textual information in the graph nodes.

\myparagraph{DiffuScene~\cite{tang2023diffuscene}.} DiffuScene is a diffusion-based retrieval method, which can be both unconditional and text-conditional. It uses a diffusion process to generate bounding boxes of an object set as the scene layout. We test its text-conditional version to perform scene synthesis on the SG-FRONT dataset. Specifically, we transfer scene graph description to sentences based on the script provided by DiffuScene. Then, we feed the sentences to the BERT~\cite{kenton2019bert} encoder to have textual embeddings for conditioning the denoising process. The experimental settings are kept the same as the original ones. DiffuScene is explicitly designed for scene completion tasks, where it uses partial inputs as a basis and generates additional content. Our task, however, concentrates on achieving fully controllable scene synthesis. This means we aim to produce scenes that precisely match the descriptions provided in the scene graph. Consequently, the evaluation of DiffuScene focuses solely on the fidelity of the generated scenes, assessing how closely the distribution of the generated content aligns with the original data. We do not enforce strict adherence to the graph constraints, recognizing that a significant portion of the content is creatively inferred or `hallucinated.'

\myparagraph{InstructScene~\cite{lin2024instructscene}.} InstructScene is another retrieval method with a closer setting to us. It stacks two diffusion-based stages: Firstly, it transfers sentences using a graph transformer denoiser to a semantic graph containing shape prior as nodes and allocates each two nodes a single edge. Secondly, another graph transformer denoiser serves as a 3D layout decoder, taking the graph and steadily denoising the 3D bounding boxes as the scene layouts. Our task focuses on graph-conditioned scene synthesis. Therefore, we conduct the experiment solely in the second stage of InstructScene. We train the graph transformer denoiser with scene graphs in SG-FRONT until its convergence. As the synthesized scenes are fully controllable by the scene graph, we are able to evaluate both scene fidelity and the performance of graph constraints. When synthesis the scenes, we retrieve the objects whose sizes are closest to the ones of generated bounding boxes. This retrieval strategy is kept the same for all methods for comparison fairness.

\subsection{Implementation Details.}

\myparagraph{Trainval and test splits.} We use the settings in DiffuScene~\cite{tang2023diffuscene} and  CommonScenes~\cite{commonscenes2023} to train and test all methods on SG-FRONT and 3D-FRONT. The whole dataset contains 4,041 bedrooms, 900 dining rooms, and 813 living rooms. The training split contains 3,879 bedrooms, 614 dining rooms, and 544 living rooms, with the rest as the test split.

\myparagraph{Batch size definition.} The two branches use individual batches in terms of the different training objectives. The layout branch uses a scene batch $B_s$ during one training step, containing all bounding boxes in scenes. 
There are two ways to determine the batch size for the shape branch.

\textit{First}, if it is the ablated version where the shape branch is without shape echoes, we can straightforwardly sample $B_o$ objects (nodes) out of the scene batch to train, as illustrated in CommonScenes~\cite{commonscenes2023}. This method allows for random sampling of objects since the lack of shape echoes means there's no requirement for the objects to originate from the same scene.

\textit{Second}, if it is the full version, we set up a maximum batch size $B_o^*$, and select scenes where the total number of objects $B_o$ closely approaches but does not exceed $B_o^*$. This method ensures that we efficiently utilize the batch capacity while adhering to the constraint of keeping the sum of objects within the predetermined maximum batch size. In this case, the batch size in shape branch $B_o$ slightly fluctuates, as the object numbers are not fixed in the scene.

\myparagraph{Training procedure.}
We train \emph{EchoLayout} (layout branch) for 2000 epochs with $B_s = 64$. The learning rate is set to [1e-4, 5e-5, 1e-5, 5e-6] at [0, 35,000, 70,000, 140,000] step. For the full \emph{EchoScene}, which integrates shape information via a shape branch into the pipeline—where both branches utilize a shared latent graph representation—we extend the training by an additional 50 epochs. The maximum batch size $B_o^*$ for the shape branch is set to 64. The learning rate is kept in the same fashion. 

\end{document}